\documentclass[11pt]{article}

\usepackage[preprint]{acl}

\usepackage{times}
\usepackage{latexsym}
\usepackage{enumitem}
\usepackage[T1]{fontenc}

\usepackage[utf8]{inputenc}

\usepackage{microtype}

\usepackage{inconsolata}

\usepackage{graphicx}
\usepackage{subcaption}
\usepackage{booktabs}
\usepackage{multirow}

\usepackage{algorithm}
\usepackage{algpseudocode}    
\usepackage{enumitem}         
\usepackage{amsthm}           

\usepackage{amsmath} 

\title{Sample Where You Struggle: Sharpening Base Model Reasoning via Entropy-Guided Power Sampling}

\author{
 \textbf{Hong Guo\textsuperscript{1}},
 \textbf{Nianhui Guo\textsuperscript{2}},
 \textbf{Christoph Meinel\textsuperscript{3}},
 \textbf{Haojin Yang\textsuperscript{2}},
\\
 \textsuperscript{1}Hasso Plattner Institute,
 \textsuperscript{2}GreenBit.AI,
 \textsuperscript{3}German University of Digital Science,
\\
 \texttt{hong.guo@hpi.de}
 \texttt{nianhui.guo@greenbit.ai}
\\
 \texttt{christoph.meinel@german-uds.de}
 \texttt{haojin.yang@greenbit.ai}
}

\begin{document}
\maketitle
\begin{abstract}
Sampling from the sequence-level power distribution elicits RL-level reasoning from base language models without any parameter updates, but the standard Metropolis--Hastings (MH), a Markov Chain Monte Carlo (MCMC) sampler, is both expensive and slow-mixing.
We trace both to a structural mismatch: the power distribution mainly departs from the base distribution at a sparse, spatially clustered set of high-entropy decision points, yet MH proposes resampling positions uniformly along the prefix---wasting compute on near-degenerate conditionals while under-mixing precisely where modes diverge. 
We propose Entropy-Guided Power Sampling (EGPS), a training-free and verifier-free approximate sampler that re-derives its proposal from token-level entropy already in the forward pass. EGPS skips deterministic blocks, localizes each MCMC move to a high-entropy neighborhood, and applies Multiple-Try Metropolis at decision points---making sampling cost scale with entropy mass rather than sequence length. EGPS targets significant speedup with accuracy that matches or improves on MH power sampling: across three base models and three benchmarks it attains best or tied-best accuracy on 5 of 9 settings; on Qwen2.5-Math-7B it is best or tied-best on all three (MATH500 $75.8\%$, HumanEval $62.2\%$, GPQA $42.4\%$), with wall-clock speedups over the MH baseline of up to $12.6\times$ across these benchmarks. Our code is available at \url{https://github.com/hongguo7/EGPS}.
\end{abstract}

\begin{figure*}[t]
  \includegraphics[width=\textwidth]{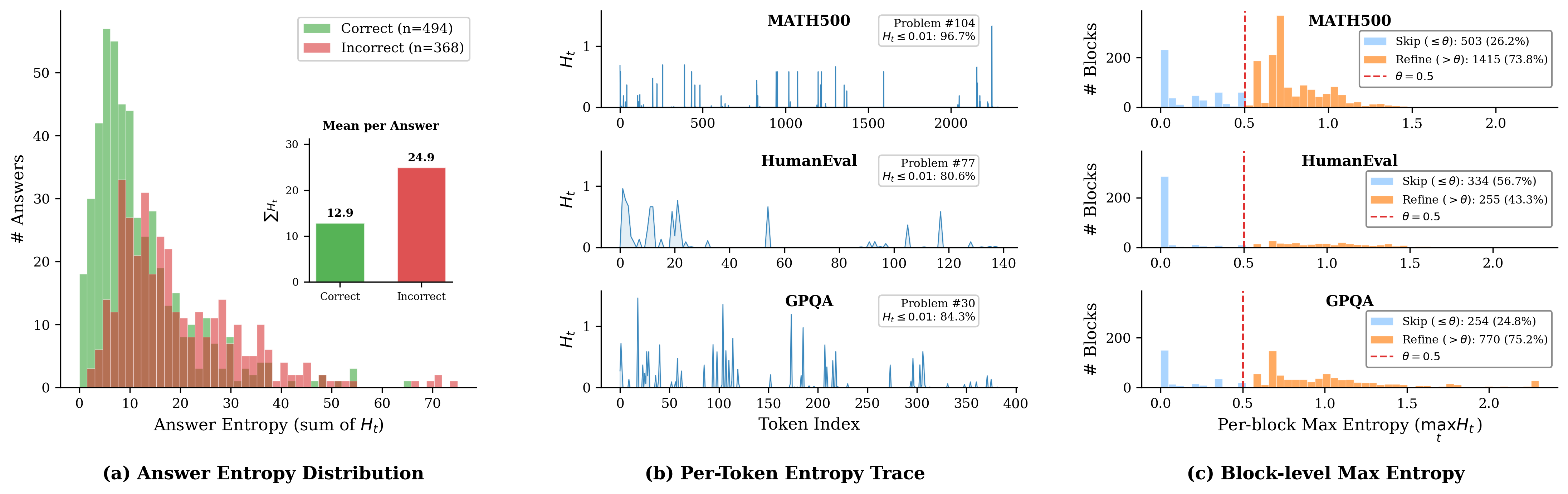}
\caption{Empirical entropy characteristics of reasoning traces. 
(a) Histogram of per-answer total entropy on merged MATH500, 
HumanEval, GPQA-Diamond, separated by correctness. 
(b) Per-token entropy $H_t$ for one selected answer per dataset. 
(c) Histogram of per-block maximum entropy $\max_t H_t$, 
shown separately for each dataset.}

  \label{fig:entropy}
\end{figure*}

\section{Introduction}
\label{sec:intro}
Reinforcement learning (RL) post-training has emerged as a common approach for enhancing the reasoning ability of large language models (LLMs), but it is costly: it typically requires carefully curated data, an elaborate training pipeline, and automated verifiers.
A growing line of work~\cite{he2025rewarding, yue2025does, song2025outcome} argues from different angles that RL post-training primarily performs distribution sharpening: rather than introducing reasoning abilities absent from the base model, it redistributes probability mass toward reasoning trajectories already favored by the base model.
This perspective motivates a natural alternative: sharpening the base model's output distribution directly at inference time, thereby bypassing the costly training process. Along this line, \citet{karan2025reasoning} propose power sampling, which uses Metropolis--Hastings (MH) to sample from the sequence-level power distribution $p^{\alpha}$ ($\alpha > 1$). Without any parameter updates, it reportedly achieves performance comparable to GRPO~\cite{shao2024deepseekmath} on benchmarks such as MATH500~\cite{lightman2024let}.


The practicality of power sampling, however, is limited by the efficiency of its MH sampler. MH generates the sequence block by block: after each block is produced, it selects a resampling start point uniformly from the entire generated prefix and regenerates from that point to the end of the current block (Fig.~\ref{fig:mtm_algo}(a)). This uniform proposal implicitly treats every token as equally worth resampling.

Yet reasoning traces are far from uniform: prior work has shown that their quality tends to be driven by a small set of critical tokens, which often exhibit high entropy~\cite{lin2024critical, wang2026beyond, yang2025less}; in particular, incorrect answers tend to have higher overall entropy than correct ones~\cite{yang2025less}, a pattern we observe empirically (Fig.~\ref{fig:entropy}a). We further characterize the entropy distribution at both the token and block levels, revealing two structural properties of high-entropy positions. First, sparsity: high-entropy positions account for only a small fraction of the sequence. Second, spatial clustering: high-entropy positions are separated by long stretches of near-deterministic tokens, forming spatially localized clusters (Fig.~\ref{fig:entropy}b). A direct consequence of these two properties is that a substantial fraction of blocks remain in a low-entropy regime throughout ($56.7\%$ on HumanEval, Fig.~\ref{fig:entropy}c). These observations reveal that MH's uniform proposal is both sample-inefficient and structurally misaligned: it over-samples confident positions while under-exploring the sparse decision points that drive quality.

Motivated by these observations, we propose Entropy-Guided Power Sampling (EGPS), which aligns the MH proposal with the entropy structure of reasoning traces. EGPS operates at two complementary granularities. At the block-level, it exploits spatial clustering through a low-entropy gate: any block whose maximum token entropy falls below a threshold $\theta$ is skipped, diverting the sampling budget away from near-deterministic regions. At the token-level, it exploits sparsity by drawing resampling points with probability proportional to token entropy, concentrating the budget on the sparse high-entropy clusters that drive quality. To further amplify exploration at these critical positions, we replace the single-proposal MH step with multiple-try Metropolis (MTM), which expands the per-step proposal pool from one to multiple candidates. Beyond the core design, we analyze the effect of the resampling search range on accuracy and provide a full implementation of EGPS within the vLLM inference framework.

Our main contributions are as follows:
\begin{itemize}[topsep=2pt, itemsep=4pt, parsep=0pt, leftmargin=2em]
    \item[(i)] We empirically characterize the entropy structure of base-model reasoning traces at both the token and block levels, identifying two structural properties of high-entropy positions: sparsity and spatial clustering. These properties point to a small set of pivotal decision points that shape reasoning quality, motivating a sampling procedure that redirects effort toward them and jointly improves inference efficiency and reasoning accuracy.
    
    \item[(ii)] We propose EGPS, a training- and verifier-free approximate power sampler for eliciting reasoning ability from base models at inference time, offering a practical alternative to RL post-training. EGPS integrates three mechanisms: entropy-triggered block skipping, entropy-weighted start-point sampling, and multiple-try Metropolis. We further analyze the effect of the resampling search range and provide a full vLLM implementation of EGPS.
    
    \item[(iii)] On three base models (Qwen2.5-Math-7B, Qwen2.5-7B, Phi-3.5-mini-instruct) and three reasoning benchmarks (MATH500, HumanEval, GPQA), EGPS achieves comparable or better accuracy than an MH Power Sampling baseline within the same vLLM framework---best or tied-best on $5$ of $9$ model--benchmark combinations; on Qwen2.5-Math-7B it is best or tied-best on all three benchmarks (MATH500 $75.8\%$, HumanEval $62.2\%$, GPQA $42.4\%$) with wall-clock speedups of up to $12.6\times$ across them.
    
\end{itemize}

\section{Related Work}
\subsection{Efficient Power Sampling}

\citet{karan2025reasoning} formalize inference-time reasoning enhancement as sampling from the sequence-level power distribution $p^\alpha$ via block Metropolis–Hastings. Their key insight is that $p^\alpha$ implicitly accounts for future path likelihoods when selecting each token, a global planning effect that token-level low-temperature sampling cannot replicate. The method matches GRPO on MATH500 without training or verifiers, but its uniform resampling scheme requires regenerating increasingly long suffixes at each block, leading to substantial computational overhead.

Follow-up work improves power sampling along two directions. On the efficiency side, \citet{ji2026scalable} show that the power distribution's conditionals decompose into a low-temperature distribution scaled by a future-quality factor, enabling rollout-based token-level approximation that bypasses iterative Markov Chain Monte Carlo (MCMC) entirely. 
\citet{azizi2026power} replace the serial MH chain with batch-parallel Sequential Monte Carlo (SMC), proving that $\tau=1/\alpha$ is the minimum-variance prefix-only proposal and shifting the bottleneck to GPU-friendly batch computation. 
\citet{abdulloh2025uncertainty} adjusts MCMC block size based on local entropy, allocating finer granularity to uncertain regions. These methods reduce per-step cost or adjust the computation allocation strategy, but computation remains distributed across all token positions. A framework-matched comparison with this method is given in Appendix~\ref{app:adaptive}.
On the sampling target side, \citet{markovic2026sampling} combine $p^\alpha$ with external reward potentials to define a reward-augmented target distribution and develop SMC-based sampling algorithms, but their reliance on an external reward model limits applicability to tasks where a reliable verifier is available. Unlike the above methods, the proposed EGPS uses only conditional entropy signals already available from the forward pass, selectively concentrating the MCMC budget on sparse positions where meaningful decision divergence exists through entropy-triggered block skipping and entropy-guided intra-block resampling, without requiring any external model. This selective allocation strategy is in principle composable with the above methods.

\subsection{Critical Tokens and Local Uncertainty in Reasoning}
The selective allocation strategy of EGPS builds on an empirical finding: reasoning difficulty concentrates at a small number of token positions. \citet{lin2024critical} identify critical tokens via large-scale rollouts and exploit this finding through token-level contrastive preference optimization (cDPO). \citet{wang2026beyond} show that high-entropy tokens act as ``forking tokens'' that steer the model toward different reasoning paths. \citet{yang2025less} find that the entropy gap between correct and incorrect trajectories is driven by a small subset of high-entropy tokens, and propose MTI to apply classifier-free guidance (CFG)~\cite{sanchez2023stay} selectively at these positions. \citet{fu2025deep} take a confidence perspective, using local log-probability to early-terminate low-quality trajectories (DeepConf).

Prior work has studied related signals---causal influence~\cite{lin2024critical}, token entropy~\cite{wang2026beyond, yang2025less}, and log-probability confidence~\cite{fu2025deep}---but only for training, decoding, or single-pass sampling. EGPS brings this signal class into MCMC power sampling, where each skipped block saves multiple MCMC iterations rather than a single forward pass.

\begin{figure*}[t]
  \includegraphics[width=\textwidth]{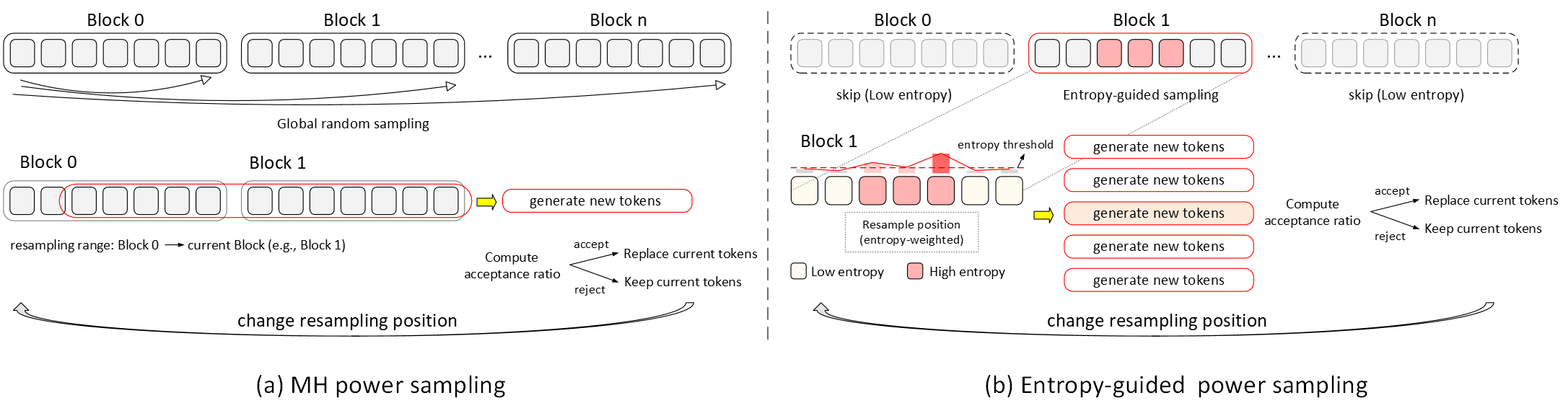}
\caption{Comparison of MH Power Sampling and EGPS. \textbf{(a)} MH Power Sampling generates the sequence block by block; after each block, it uniformly selects a start point from the generated prefix and regenerates from there to the block's end. \textbf{(b)} EGPS replaces the uniform proposal with a two-granularity, entropy-guided one: a block-level low-entropy gate skips near-deterministic blocks, and a token-level entropy-weighted distribution concentrates start points on high-entropy positions. Multiple-try Metropolis (MTM) then expands the per-step proposal pool from one to multiple candidates.}
  \label{fig:mtm_algo}
\end{figure*}

\section{Method}\label{sec:method}
\subsection{Preliminaries}
\label{sec:prelim}
\paragraph{Power distributions and MH sampling.}
Let~$\mathcal{X}$ be a finite token vocabulary and~$p$ an
autoregressive language model that factorizes as
\begin{equation}\label{eq:joint}
  p(x_{0:T}) = \prod_{t=0}^{T} p(x_t \mid x_{<t}),
\end{equation}
where $x_t \in \mathcal{X}$ denotes the token at position~$t$,
$x_{<t} = (x_0, \dots, x_{t-1})$ its prefix,
and $T$ the sequence length.
The power distribution~$p^\alpha$ with sharpening
exponent~$\alpha > 1$ assigns each sequence a probability
proportional to
\begin{equation}\label{eq:power-dist}
  p^\alpha(x_{0:T}) \;\propto\; p(x_{0:T})^\alpha.
\end{equation}
We write $\tilde{p}(x_{0:T}) := p(x_{0:T})^{\alpha}$ for this
unnormalized target density, so that
$p^{\alpha} = \tilde{p}/Z$ with normalizer
$Z = \sum_{\mathbf{x} \in \mathcal{X}^{T}} \tilde{p}(\mathbf{x})$;
throughout, $p$ denotes the normalized base-model distribution
and $\tilde{p}$ the unnormalized target.
A larger~$\alpha$ concentrates mass on high-likelihood sequences;
$\alpha = 1$ recovers the base model.
Sampling from~$p^\alpha$ is not equivalent to
low-temperature sampling ($\tau = 1/\alpha$), which exponentiates
each conditional $p(x_t \mid x_{<t})$ independently rather than
accounting for the exponentiated joint likelihood of all future
continuations.

While the unnormalized density~$\tilde{p}(\mathbf{x})$ is readily
available for any sequence~$\mathbf{x} = x_{0:T}$, the
normalizer~$Z$ requires summing over all of~$\mathcal{X}^T$, which
is computationally intractable.
The MH algorithm circumvents this by
constructing a Markov chain whose stationary distribution
is~$p^\alpha$: given a current sequence~$\mathbf{x}$, a
candidate~$\mathbf{x}'$ drawn from a proposal distribution~$q$
is accepted with probability
\begin{equation}\label{eq:mh-acceptance}
  A(\mathbf{x}', \mathbf{x})
  = \min\!\left(1,\;
    \frac{\tilde{p}(\mathbf{x}') \cdot q(\mathbf{x} \mid \mathbf{x}')}
         {\tilde{p}(\mathbf{x}) \cdot q(\mathbf{x}' \mid \mathbf{x})}
  \right),
\end{equation}
where $\tilde{p}(\mathbf{x})$ and~$\tilde{p}(\mathbf{x}')$ are the
unnormalized target densities of the current and candidate
sequences, and $q(\mathbf{x}' \mid \mathbf{x})$,
$q(\mathbf{x} \mid \mathbf{x}')$ are the forward and reverse
proposal probabilities.
The normalizer~$Z$ cancels in the ratio,
so only log-probabilities under~$p$ are needed.
Following \citet{karan2025reasoning}, we instantiate the
proposal by low-temperature resampling: a resampling position
is chosen, and the suffix is regenerated tokenwise from
$q = p_{\tau}$ with temperature $\tau = 1/\alpha$, i.e., each
conditional $p(x_t \mid x_{<t})$ tempered at $\tau$.
The MH correction in Eq.~\eqref{eq:mh-acceptance} accounts for
the gap between this token-level proposal and the
sequence-level target~$p^{\alpha}$.

\paragraph{Conditional entropy.}
The conditional entropy at position~$t$ is
\begin{equation}\label{eq:entropy}
  H_t = -\sum_{v \in \mathcal{X}}
    p(v \mid x_{<t}) \log p(v \mid x_{<t}),
\end{equation}
where $p(v \mid x_{<t})$ is the model's next-token distribution
given prefix~$x_{<t}$.
$H_t$ quantifies the model's uncertainty at position~$t$ and
can be computed directly from the logits produced during
generation at no extra cost.

\begin{algorithm*}[t]
\caption{Entropy-Guided Power Sampling (EGPS)}
\label{alg:EGPS}
\begin{algorithmic}[1]
\Statex \textbf{Input:} base $p$; proposal $q = p_{\tau}$, $\tau{=}1/\alpha$ (low-temp.); power $\alpha$; length $T$
\Statex \textbf{Hyperparams:} block size $B$; MCMC steps $N_{\mathrm{MCMC}}$; 
        proposals $K$; entropy threshold $\theta$; runway $\delta$; 
        candidate range $S$
\Statex \textbf{Output:} $(x_1, \ldots, x_T) \sim p^\alpha$

\For{$k \leftarrow 0$ \textbf{to} $\lceil T/B \rceil - 1$}

  \State $x_t \sim q(\cdot \mid x_{<t})$ for $t \in [kB{+}1, (k{+}1)B]$; compute $H_t$
  
  \If{$\max_t H_t \leq \theta$ for $t \in [kB{+}1, (k{+}1)B]$}
    \State \textbf{continue} \Comment{skip low-entropy block}
  \EndIf

  \For{$n \leftarrow 1$ \textbf{to} $N_{\mathrm{MCMC}}$}

    \State Identify high-entropy positions $\{t \in S : H_t > \theta\}$ and 
           extend by $\delta$ tokens leftward to form candidate set $C$
    \State Sample $m \in C$ with weight $\propto \max_{t' \in [m, m+\delta]} H_{t'}$
    
    \State $\mathbf{x}'_j \sim q$ for $j = 1, \ldots, K$, resampling $[m, (k{+}1)B]$
    \State Compute $\{w_j\}$ (Eq.~\ref{eq:mtm-weight});  $W \leftarrow \sum_j w_j$
    \State $j^* \sim \mathrm{Categorical}(w_1/W, \ldots, w_K/W)$
    
    \State $u \sim \mathrm{Uniform}(0, 1)$
    \State \textbf{if} $u \leq A_{\mathrm{MTM}}$ (Eq.~\ref{eq:mtm-acceptance}) \textbf{then} 
           $x_{m:(k+1)B} \leftarrow \mathbf{x}'_{j^*}$
  
  \EndFor
\EndFor
\State \Return $(x_1, \ldots, x_T)$

\end{algorithmic}
\end{algorithm*}

\subsection{Entropy-Guided Power Sampling}
\label{sec:egps}
Building on the two structural properties identified in
\S\ref{sec:intro}, namely the sparsity and spatial clustering of high-entropy positions, MH power
sampling spends most of its computation on deterministic, low-entropy
regions. As shown in Fig.~\ref{fig:mtm_algo}, EGPS uses entropy to concentrate the MCMC budget where decisions actually occur, and upgrades the single-proposal MH step to Multiple-Try Metropolis (MTM) to widen the per-step search.
This section describes the algorithm (Algorithm~\ref{alg:EGPS}) and analyzes
its computational cost.

\paragraph{Generation and triggering.}
EGPS partitions the target sequence into blocks of size $B$,
indexed by $k = 0, 1, \ldots, \lceil T/B \rceil - 1$ (block $k$
covers positions $[kB{+}1,\,(k{+}1)B]$), and generates them
sequentially. For each block, EGPS enters the MCMC
refinement loop only when $\max_t H_t > \theta$ within the block;
otherwise the block is skipped. Low-entropy blocks rarely contain
decision points, so skipping them saves sampling budget.

\paragraph{Position selection.}
Once a block has triggered refinement, every MCMC step inside the
loop must choose a resampling start $m$; EGPS does so in two
stages. First, it fixes a search range
$S \subseteq [1,\,(k{+}1)B]$, whose two extremes are
$S = [kB{+}1,\,(k{+}1)B]$ (local) and
$S = [1,\,(k{+}1)B]$ (global). The local range restricts
each move to the current block and is cheaper ($O(KT)$; see the
cost analysis below), suiting errors confined to a block; the
global range can also revise earlier blocks ($O(KT^{2}/B)$ in
the worst case), suiting errors that propagate across
blocks---empirically, global is stronger on math and science,
local on code (\S\ref{sec:main_results}).
Second, within $S$, EGPS restricts the candidates to the
high-entropy set $\{t \in S : H_t > \theta\}$ and extends it
leftward by $\delta$ tokens to form the runway region,
from which $m$ is sampled with probability proportional to
\begin{equation}\label{eq:pos-weight}
w_{\text{pos}}(m \mid \mathbf{x}) \;\propto\; \max_{t \in [m,\,m+\delta]} H_t(\mathbf{x}).
\end{equation}
This concentrates $m$ just before the high-entropy tokens: backing
up $\delta$ tokens lets the model rephrase the lead-in and traverse
the decision point along a different path, whereas resampling at
the decision point itself is more likely to reproduce similar
hesitations.

\paragraph{Multiple-try proposals and weighted selection.}
At each MCMC step, EGPS generates $K$ candidate suffixes
$\mathbf{x}'_1, \ldots, \mathbf{x}'_K$ in parallel from the proposal
distribution $q$, where $K \geq 1$ is the number of candidates per
step. Each $\mathbf{x}'_j$ shares the same prefix as the current
state $\mathbf{x}$ and differs only on the suffix from position
$m$ onward.

To pick among the $K$ candidates, EGPS assigns each candidate an importance weight $w_j$, defined as the Metropolis--Hastings
ratio of candidate $j$ against the current state under the target
$p^{\alpha}$:
\begin{equation}
w_j \;=\; \frac{\tilde{p}(\mathbf{x}'_j)\, q(\mathbf{x} \mid \mathbf{x}'_j)}
                 {\tilde{p}(\mathbf{x})\, q(\mathbf{x}'_j \mid \mathbf{x})}.
\label{eq:mtm-weight}
\end{equation}
A candidate $\mathbf{x}'_{j^{*}}$ is then drawn with probability
proportional to $w_j$ (here $j^{*}$ denotes its index), and the
move to it is accepted with probability
\begin{equation}
A_{\mathrm{MTM}} \;=\; \min\!\Big(1,\; \frac{W}{W - w_{j^{*}} + 1}\Big),
\label{eq:mtm-acceptance}
\end{equation}
where $W = \sum_j w_j$ is the total weight summed over all $K$
candidates.
This multiple-proposal mechanism is the independent
multiple-try Metropolis (I-MTM) scheme of
\citet{martino2018review}: Eqs.~\eqref{eq:mtm-weight}
and~\eqref{eq:mtm-acceptance} are the standard I-MTM weight and
acceptance ratio (Eqs.~18--19 therein), expressed with each
weight normalized by the current state, so the $+1$ in the
denominator is the normalized current-state term rather than an
added constant (derivation in Appendix~\ref{app:imtm}).
Unlike standard MTM, I-MTM requires no $K{-}1$ auxiliary
reference draws to balance the current state, making it simpler
and computationally cheaper.
At $K = 1$, Eq.~\eqref{eq:mtm-acceptance} reduces to
the standard MH acceptance $\min(1, w_1)$.

\paragraph{Sources of approximation.}
Two of the mechanisms above make the EGPS kernel a controlled approximation rather than an exact
$p^{\alpha}$-invariant sampler. First, since the resampling
position is drawn non-uniformly
($m \sim w_{\text{pos}}(\cdot \mid \mathbf{x})$,
Eq.~\eqref{eq:pos-weight}), the forward and reverse position
probabilities no longer cancel: the exact acceptance ratio
carries an additional position factor
$w_{\text{pos}}(m \mid \mathbf{x}') / w_{\text{pos}}(m \mid \mathbf{x})$
multiplying Eq.~\eqref{eq:mh-acceptance}, and in MTM each weight
$w_j$ carries its own such factor; EGPS omits these factors.
Second, entropy-triggered skipping removes MCMC steps on
low-entropy blocks, where the conditionals are already
near-deterministic and refinement would change essentially no
tokens.
Both deviations are empirically negligible: restoring the
position factor shifts the log acceptance ratio by an amount
concentrated tightly at zero with no systematic direction, and a
Kolmogorov--Smirnov test on per-sequence mean log-likelihood
cannot distinguish end-to-end EGPS (skips included) from MH power sampling for the local variants, while the global
variant shifts toward higher likelihood---a task-favorable
sharpening consistent with its stronger downstream accuracy.
Derivations and measurements are in
Appendix~\ref{app:exactness}.

\paragraph{Computational cost.}
The original MH sampler regenerates on average~$kB/2$ tokens at
the $k$-th block, giving a total of
$E_{\mathrm{MH}} \approx N_{\mathrm{MCMC}} \cdot T^{2}/4B$,
quadratic in~$T$ (derivation in Appendix~\ref{app:complexity}). In local mode, EGPS restricts each rewriting
step to the current block (at most~$B$ tokens), yielding
\begin{equation}\label{eq:cost-local}
  E_{\text{local}}
  \leq (1 - s) \cdot K \cdot N_{\mathrm{MCMC}} \cdot T,
\end{equation}
where $s$ is the block skip rate. The scaling reduces from
$O(T^{2}/B)$ to~$O(KT)$. In global mode the worst-case scaling
remains~$O(KT^{2}/B)$, but entropy weighting concentrates~$m$
near the triggering block, so the empirical cost is substantially
lower (see~\S\ref{sec:ablation}).

\section{Experiment}
\label{sec:experiment}
\subsection{Experimental Setup}
\label{sec:setup}


\begin{table*}[t]
\centering

\footnotesize  
\caption{%
  Main results (pass@1) on MATH500, HumanEval, and GPQA.
  Higher is better; bold marks the best non-RL result per
  (model, benchmark). For EGPS, we report the best accuracy over swept
  thresholds $\theta \in \{0.1, 0.2, 0.4, 0.8\}$; per-$\theta$ accuracy
  and wall-clock speedup are in §\ref{sec:ablation}. GRPO (italic) is
  shown as a reference upper bound from RL post-training and is not
  compared against.
  Baseline accuracies are cited from their source papers
  (\S\ref{sec:setup}).
}

\label{tab:main}
\begin{tabular}{@{}l ccc@{}}
\toprule
\textbf{Method} & \textbf{MATH500} & \textbf{HumanEval} & \textbf{GPQA} \\
\midrule

\multicolumn{4}{l}{\textbf{Qwen2.5-Math-7B}} \\
Base \citep{karan2025reasoning}                             & 0.496 & 0.329 & 0.278 \\
Low-temp \citep{karan2025reasoning}                         & 0.690 & 0.512 & 0.353 \\
Best-of-$N$ ($N{=}32$) \citep{ji2026scalable}           & 0.684 & 0.512 & 0.343 \\
MH Power Sampling \citep{karan2025reasoning}               & 0.748 & 0.573 & 0.389 \\
Scalable PS \citep{ji2026scalable}                        & \textbf{0.758} & 0.604 & 0.409 \\
Power-SMC ~\citep{azizi2026power}                        & 0.742 & 0.585 & 0.349 \\
APPS (p-only) ~\citep{nguyen2026model}                     & 0.744 & \textbf{0.622} & 0.323 \\
\textbf{EGPS\textsubscript{G}} (MTM)   & \textbf{0.758} & 0.604  &\textbf{0.424}  \\ 
\textbf{EGPS\textsubscript{L}} (MTM)   & 0.754 & \textbf{0.622} &0.384  \\
\cmidrule{1-4}

GRPO (MATH) \citep{karan2025reasoning}                     & \textit{0.785} & \textit{0.537} & \textit{0.399} \\
\midrule

\multicolumn{4}{l}{\textbf{Qwen2.5-7B}} \\
Base \citep{karan2025reasoning}                            & 0.498 & 0.329 & 0.278 \\
Low-temp \citep{karan2025reasoning}                          & 0.628 & 0.524 & 0.303 \\
Best-of-$N$ ($N{=}32$) \citep{ji2026scalable}            & 0.650 & 0.609 & 0.282 \\
MH Power Sampling \citep{karan2025reasoning}                & 0.706 & 0.622 & 0.318 \\
Scalable PS \citep{ji2026scalable}                        & 0.708 & \textbf{0.756} & 0.349 \\
Power-SMC ~\citep{azizi2026power}                        & 0.716 & 0.683 & 0.313 \\
APPS (p-only) ~\citep{nguyen2026model}                     & \textbf{0.726} & 0.561 & 0.354 \\
\textbf{EGPS\textsubscript{G}} (MTM)   & 0.710 & 0.610  &0.354 \\
\textbf{EGPS\textsubscript{L}} (MTM)   & 0.712 & 0.646 & \textbf{0.359} \\
\cmidrule{1-4}
GRPO (MATH) \citep{karan2025reasoning}                      & \textit{0.740} & \textit{0.561} & \textit{0.354} \\

\midrule

\multicolumn{4}{l}{\textbf{Phi-3.5-mini-instruct}} \\
Base \citep{karan2025reasoning}                           & 0.400 & 0.213 & 0.273 \\
Low-temp \citep{karan2025reasoning}                          & 0.478 & 0.585 & 0.293 \\
MH Power Sampling \citep{karan2025reasoning}               & 0.508 & \textbf{0.732} & \textbf{0.364} \\
\textbf{EGPS\textsubscript{G}} (MTM)   & \textbf{0.520} &0.677  &0.343 \\
\textbf{EGPS\textsubscript{L} }(MTM)   & 0.504 & 0.695 &0.338 \\
\cmidrule{1-4}

GRPO (MATH) \citep{karan2025reasoning}                    & \textit{0.406} &\textit{0.134} & \textit{0.359} \\
\bottomrule
\end{tabular}

\end{table*}

\paragraph{Datasets and models.}
We evaluate on three reasoning benchmarks covering distinct domains---MATH500~\citep{lightman2024let} (mathematical reasoning), HumanEval~\citep{chen2021evaluating} (code generation), and GPQA Diamond~\citep{rein2023gpqa} (graduate-level scientific question answering)---using three publicly available models that span distinct training objectives and capacities: Qwen2.5-Math-7B~\citep{yang2024qwen2technicalreport} (a math-specialized base model), Qwen2.5-7B~\citep{yang2024qwen2technicalreport} (a general-purpose base model), and Phi-3.5-mini-instruct (a small-scale instruction-tuned model). All methods are evaluated under a single-shot protocol (one response per question), ensuring fair comparison and reflecting realistic deployment.

\paragraph{Baselines.}
We compare against seven representative methods. For all baselines, we cite the accuracy numbers reported in their papers: Base ($\tau{=}1.0$) and Low-temperature sampling ($\tau{=}0.25$), with numbers taken from \citet{karan2025reasoning}; Best-of-$N$ ($N{=}32$), with numbers taken from \citet{ji2026scalable}; MH Power Sampling~\citep{karan2025reasoning}, Scalable Power Sampling~\citep{ji2026scalable}, Power-SMC~\citep{azizi2026power}, and APPS (the reward-free p-only variant)~\citep{nguyen2026model}.

\paragraph{EGPS configurations.}
We evaluate EGPS in two configurations:
\textbf{EGPS\textsubscript{G}}: the resample range spans all generated tokens, exploring the upper bound of accuracy.
\textbf{EGPS\textsubscript{L}}: the resample range is confined to the current block, exploring the upper bound of efficiency. Each MCMC step uses an MTM proposal with $K{=}5$ candidate sequences.

\paragraph{Implementation and speedup measurement.}
To ensure comparability with published baselines, we adopt the same hyperparameters as \citet{karan2025reasoning}: block size $B{=}192$, power exponent $\alpha{=}4$, proposal temperature $\tau{=}0.25$, $N_{\mathrm{MCMC}}{=}10$ MCMC steps per refinement block, and maximum generation length $T_{\max}{=}3072$. The entropy threshold $\theta$ is selected via ablation (see~\S\ref{sec:ablation}). All experiments are conducted on the NVIDIA A100-40G GPU, with EGPS and all vLLM-based baselines implemented on vLLM 0.11.0 (PyTorch 2.8.0, CUDA 12.8, Transformers 4.55.2, Python 3.10). Since vLLM accelerates inference for any algorithm through KV-cache and batched scheduling optimizations, comparing EGPS (on vLLM) against published speedup numbers from non-vLLM implementations would conflate algorithmic with framework-level gains. We therefore re-implement MH Power Sampling on vLLM---preserving its original algorithm (uniform-position resampling with single-proposal MH)---and report all speedups against this vLLM-based MH Power Sampling baseline; accuracy numbers, in contrast, are taken directly from the original papers. Prompt templates are listed in Appendix~\ref{app:prompts}.

\begin{figure*}[htbp]
  \centering
    \centering
    \includegraphics[width=1\textwidth]{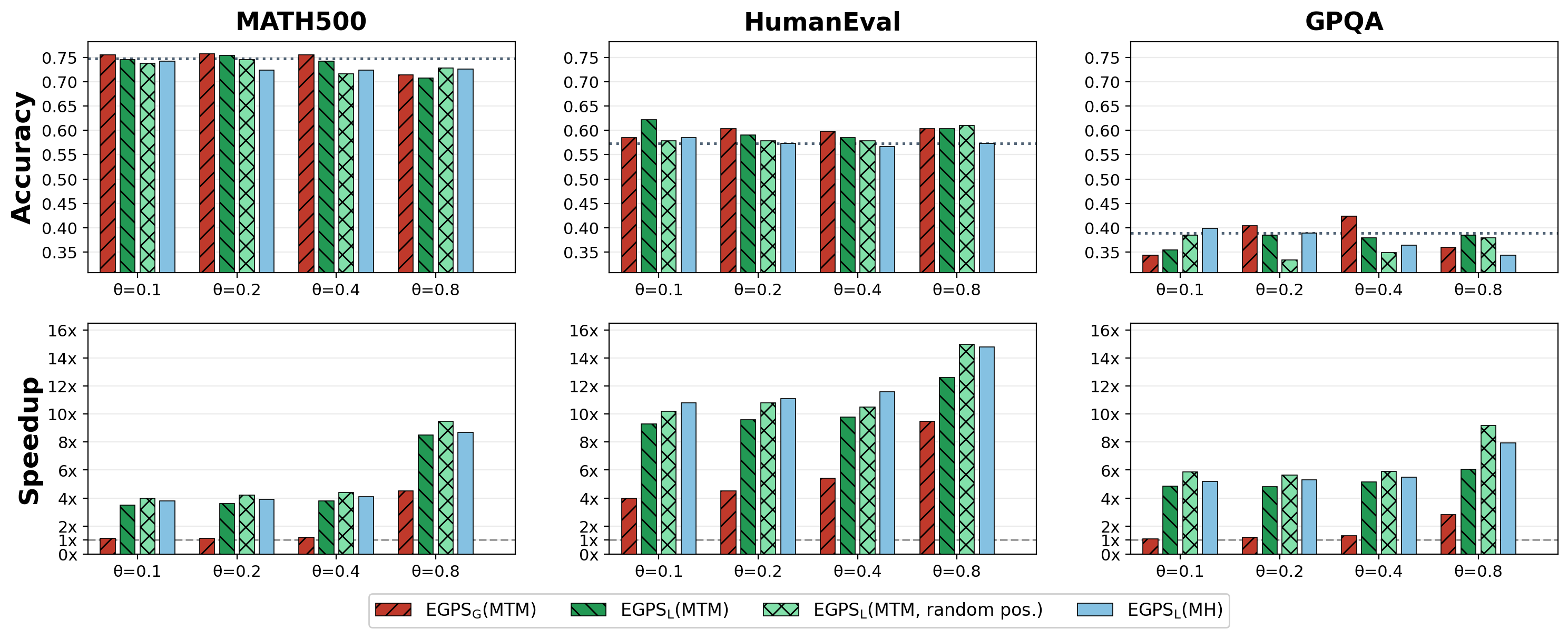}
    \caption{Ablation on Qwen2.5-Math-7B, sweeping $\theta \in \{0.1, 0.2, 0.4, 0.8\}$ on MATH500, HumanEval, and GPQA. Top row: pass@1 accuracy (dotted line: vLLM-based MH Power Sampling baseline). Bottom row: wall-clock speedup over the same baseline (dashed line at $1\times$). EGPS$_\mathrm{G}$(MTM) and EGPS$_\mathrm{L}$(MTM) use Global and Local resample range; EGPS$_\mathrm{L}$(MTM, random pos.) replaces entropy-guided position sampling with random sampling; EGPS$_\mathrm{L}$(MH) replaces MTM with single-proposal MH.}
\label{fig:ablation}
    \label{fig:sub1}
\end{figure*}

\subsection{Main Results}
\label{sec:main_results}

Table~\ref{tab:main} reports pass@1 accuracy and Figure~\ref{fig:ablation} (bottom row) reports wall-clock speedup on Qwen2.5-Math-7B relative to a same-framework vLLM-based MH Power Sampling (MH PS)~\citep{karan2025reasoning} baseline. Positioned for significant speedup at comparable-or-better accuracy, EGPS matches or surpasses reward-free SOTA on most settings while delivering up to a $12.6\times$ speedup; settings where multi-sample baselines lead are analyzed below.

EGPS improves over MH PS on most model--benchmark combinations, with the smallest gains on the most training-aligned task and the largest on broader reasoning: on Qwen2.5-Math-7B the absolute accuracy gains are $+1.0\%$ (MATH500), $+4.9\%$ (HumanEval), $+3.5\%$ (GPQA); on Qwen2.5-7B they are $+0.6\%$, $+2.4\%$, $+4.1\%$. This matches our central premise: when output quality is bottlenecked by a small number of critical decisions, targeting those decisions yields larger returns than adding generic sampling effort.

The pattern also holds against more compute-hungry baselines: Best-of-$N$~\citep{ji2026scalable} ($N{=}32$ generations), Scalable Power Sampling (Scalable PS)~\citep{ji2026scalable}, Power-SMC~\citep{azizi2026power} (multi-chain or particle-based MCMC), and APPS~\citep{nguyen2026model}, each using substantially more sampling than EGPS's single chain over entropy-skipped blocks. EGPS matches or exceeds them on most settings: on Qwen2.5-Math-7B it ties Scalable PS on MATH500 ($75.8\%$) and APPS on HumanEval ($62.2\%$) and surpasses both on GPQA ($42.4\%$ vs.\ $40.9\%/32.3\%$); on Qwen2.5-7B, EGPS\textsubscript{L} surpasses Power-SMC on GPQA ($35.9\%$ vs.\ $31.3\%$) and Best-of-$N$ on MATH500 by $+6.2\%$. Two settings favor multi-sample methods: Scalable PS leads EGPS\textsubscript{L} by $+11.0\%$ on Qwen2.5-7B HumanEval, and on Phi-3.5-mini-instruct MH PS retains the lead on HumanEval and GPQA, which we attribute to a sharper output distribution at small instruction-tuned scale that leaves less high-entropy mass to redirect.

The G/L contrast is task-dependent. EGPS\textsubscript{G} outperforms EGPS\textsubscript{L} on math and science reasoning (Qwen-Math GPQA $42.4\%$ vs.\ $38.4\%$; Phi-3.5 MATH500 $52.0\%$ vs.\ $50.4\%$), consistent with errors propagating across blocks; the ranking reverses on code generation, where EGPS\textsubscript{L} wins (Qwen-Math HumanEval $62.2\%$ vs.\ $60.4\%$; Qwen-7B HumanEval $64.6\%$ vs.\ $61.0\%$), consistent with errors localized within the current block. The two configurations suit different error structures rather than acting as a single hyperparameter.

On efficiency, EGPS\textsubscript{L} reaches $3.5\text{--}8.5\times$ on MATH500, $9.3\text{--}12.6\times$ on HumanEval, and $4.8\text{--}6.1\times$ on GPQA on Qwen2.5-Math-7B, with EGPS\textsubscript{G} reaching $1.1\text{--}4.5\times$, $4.0\text{--}9.5\times$, $1.1\text{--}2.8\times$ respectively. Because the baseline runs in the same framework (vLLM), these are algorithmic gains, attributable to entropy-triggered block skipping (avoiding refinement on near-deterministic blocks) and entropy-weighted position sampling (focusing MCMC where correction is needed). Hardware-independent, mechanism-level efficiency metrics---forward-pass counts, skip statistics, and peak memory---are in Appendix~\ref{app:efficiency}.

Sampling-based methods (MH PS, Scalable PS, Power-SMC, APPS, and EGPS) avoid the pass@$k$ and cross-task degradation associated with RL-based distribution sharpening~\citep{yue2025does}: GRPO~\citep{karan2025reasoning} trained on MATH reaches $78.5\%$ on Qwen2.5-Math-7B MATH500 but collapses on Phi-3.5 ($40.6\%$ MATH500, $13.4\%$ HumanEval), while these methods hold stable performance bands within each base model. EGPS matches the strongest sampling baselines with a single MCMC chain, and on Qwen2.5-Math-7B surpasses GRPO on HumanEval ($+8.5\%$) and GPQA ($+2.5\%$); EGPS\textsubscript{L} also exceeds GRPO on Qwen2.5-7B GPQA ($35.9\%$ vs.\ $35.4\%$). EGPS\textsubscript{G} and EGPS\textsubscript{L} span a deployment curve along accuracy and efficiency.

Table~\ref{tab:main} follows the single-seed best-accuracy protocol of \citet{karan2025reasoning}. To verify robustness, we additionally rerun EGPS and our reproduced MH PS baseline over 8 seeds with two-sided paired $t$-tests: within $\theta \le 0.4$, the MATH500 gains are significant for both variants at the $5\%$ level, while HumanEval and GPQA remain on par at large speedups; full multi-seed tables and analysis are in Appendix~\ref{app:significance}.

\subsection{Ablation Study}
\label{sec:ablation}

We ablate EGPS on Qwen2.5-Math-7B by sweeping $\theta \in \{0.1, 0.2, 0.4, 0.8\}$ across four variants (Figure~\ref{fig:ablation}; full numerical results in Appendix~\ref{app:theta-numbers}): EGPS\textsubscript{G}(MTM) and EGPS\textsubscript{L}(MTM) are the full configurations with Global and Local resample range; EGPS\textsubscript{L}(MTM, random pos.) replaces entropy-guided position sampling with uniform random sampling; and EGPS\textsubscript{L}(MH) replaces multiple-try Metropolis with single-proposal MH. The sweep isolates two design components (entropy-guided position sampling and MTM) and one tunable hyperparameter ($\theta$).

Replacing entropy-guided position sampling with uniform random sampling causes the largest damage on the hardest reasoning task. The gap between EGPS\textsubscript{L}(MTM) and EGPS\textsubscript{L}(MTM, random pos.) is small on MATH500 and HumanEval across $\theta$, but widens sharply on GPQA, reaching $+5.1\%$ at $\theta=0.2$ ($38.4\%$ vs.\ $33.3\%$). This mirrors the cross-model pattern observed in the main results: when reasoning quality is bottlenecked by a small number of high-entropy decisions, uniform sampling spends MCMC budget on positions where the base model is already confident, while the entropy-weighted variant concentrates effort exactly where it matters. On tasks where most positions either matter or are already deterministic, the choice of sampling distribution carries less weight.

Replacing MTM with single-proposal MH causes consistent but more uniform accuracy drops. EGPS\textsubscript{L}(MTM) outperforms EGPS\textsubscript{L}(MH) by $+3.0\%$ on MATH500 at $\theta=0.2$ ($75.4\%$ vs.\ $72.4\%$) and by comparable margins elsewhere, with the gap holding broadly across $\theta$ rather than scaling with task difficulty. This makes MTM look less like a fix targeting any specific failure mode and more like a generic per-step boost from broader proposal diversity. The two components are therefore complementary: entropy-weighted position sampling routes MCMC effort to the right positions, while MTM improves the proposal made at each chosen position.

The threshold $\theta$ exposes an accuracy--efficiency knob with a favourable curve. Speedup grows roughly monotonically with $\theta$ across all benchmarks (e.g.\ EGPS\textsubscript{L} on MATH500: $3.5\times$ at $\theta=0.1$ rising to $8.5\times$ at $\theta=0.8$), while accuracy is stable and insensitive to the exact value within $\theta \in [0.1, 0.4]$. Higher $\theta$ skips more low-entropy blocks, so EGPS increasingly trusts the base model on near-deterministic regions; the curve degrades once $\theta$ is high enough to skip blocks that contain critical decisions. Where this happens depends on the dataset's block-entropy distribution (Fig.~\ref{fig:entropy}c): on MATH500, whose per-block entropy mass peaks at $0.6$--$0.75$, accuracy drops systematically at $\theta = 0.8$ (confirmed over 8 seeds, Appendix~\ref{app:significance}), whereas HumanEval and GPQA peak outside this range and show no systematic drop---single-seed dips at high $\theta$ are within seed noise. In practice we recommend $\theta \in [0.1, 0.4]$ as a robust default requiring no per-benchmark tuning---lower $\theta$ favors accuracy, higher favors speed---picked once without model surgery or retraining.

Taken together, these ablations decompose EGPS's gains into separable contributions: position sampling drives the task-specific accuracy improvements seen in the main results, MTM provides a stable per-step boost across tasks, and $\theta$ exposes a smooth deployment curve between accuracy-favoring and efficiency-favoring operating points. Removing either component degrades accuracy in measurably different patterns, indicating the two are non-redundant. Further component-isolating ablations---proposal count $K$, MCMC depth, runway, and selection rule---are in Appendix~\ref{app:ablation-full}.
\section{Conclusion}
\label{sec:conclusion}

In this paper, we presented Entropy-Guided Power Sampling (EGPS), a 
training-free and verifier-free sampler for eliciting RL-level reasoning 
from base language models. EGPS re-derives its proposal from the 
token-level entropy already produced in the forward pass: it skips 
deterministic blocks, localizes each Metropolis--Hastings move to a 
high-entropy neighborhood, and applies Multiple-Try Metropolis at 
decision points, so sampling cost scales with entropy mass rather than 
sequence length. Experiments on three base models and three reasoning 
benchmarks show that EGPS attains the best or tied-best accuracy on 5 of 
9 settings and yields up to a $12.6\times$ wall-clock speedup over the MH 
baseline on Qwen2.5-Math-7B; 
a token-level analysis confirms the gain stems from the proposal 
design rather than the inference framework. 
We hope these findings 
encourage further study of structurally-aware, inference-time alternatives 
to RL post-training. 
\section*{Limitations}

Our experiments cover models up to 7B parameters on three reasoning benchmarks spanning mathematics, code, and science. The two structural properties EGPS exploits---sparsity and spatial clustering of high-entropy positions (Fig.~\ref{fig:entropy})---are validated in this regime; whether they persist at 70B+ scales, or under stronger instruction tuning, which sharpens the output distribution and leaves less redirectable entropy mass (\S\ref{sec:main_results}), remains open. In non-reasoning regimes such as open-ended generation and dialogue, high entropy often reflects stylistic freedom rather than decision points, so the motivation itself may weaken. EGPS exposes a single entropy threshold $\theta$; accuracy is insensitive to its exact value within the recommended range $[0.1, 0.4]$ (\S\ref{sec:ablation}), but the optimal value depends on the dataset's block-entropy distribution, and automatic selection (e.g.\ calibration from a small held-out set or online estimation from the running entropy distribution) remains future work. Finally, we do not formally characterize how the entropy-guided proposal affects the mixing of the power distribution target $p^\alpha$; a mixing-time analysis would clarify when the observed gains should be expected to hold.

\section*{Acknowledgments}

The computing resources underlying this work were funded by the Federal
Ministry of Research, Technology and Space under the funding code
``KI-Servicezentrum Berlin-Brandenburg'' 16IS22092. Responsibility for
the content of this publication remains with the author.

\bibliography{custom}

\clearpage
\appendix
\section*{Appendix}
\setcounter{topnumber}{4}
\setcounter{totalnumber}{6}
\setcounter{dbltopnumber}{4}
\renewcommand{\topfraction}{0.99}
\renewcommand{\textfraction}{0.01}
\renewcommand{\floatpagefraction}{0.8}
\renewcommand{\dbltopfraction}{0.99}
\renewcommand{\dblfloatpagefraction}{0.8}

\section{Exactness of the EGPS Kernel}
\label{app:exactness}

This appendix derives the exact acceptance ratios that a fully
$p^{\alpha}$-invariant version of EGPS would require, identifies
the factor our implementation omits, and measures the empirical
effect of the omission together with that of entropy-triggered
skipping.

\paragraph{Exact acceptance ratios.}

In EGPS the resampling position is drawn non-uniformly,
$m \sim w_{\text{pos}}(\cdot \mid \mathbf{x})$
(Eq.~\ref{eq:pos-weight}), so the forward and reverse position
probabilities differ and no longer cancel in the MH ratio. 
For $K=1$, the exact acceptance probability is 
\begin{equation}\label{eq:mh-exact}
\resizebox{.9\hsize}{!}{$\displaystyle
A_{\text{full}} = \min\!\left(1,\;
\frac{\tilde{p}(\mathbf{x}')\, q(\mathbf{x} \mid \mathbf{x}')}
     {\tilde{p}(\mathbf{x})\, q(\mathbf{x}' \mid \mathbf{x})}
\cdot
\frac{w_{\text{pos}}(m \mid \mathbf{x}')}
     {w_{\text{pos}}(m \mid \mathbf{x})}
\right)$},
\end{equation}
whose first factor is the ratio of Eq.~\eqref{eq:mh-acceptance} and whose second factor is the position factor. 
For $K>1$
(I-MTM), each candidate weight carries its own factor,
\begin{equation}\label{eq:mtm-exact}
w_j^{\text{full}} =
\frac{\tilde{p}(\mathbf{x}'_j)\, q(\mathbf{x} \mid \mathbf{x}'_j)}
     {\tilde{p}(\mathbf{x})\, q(\mathbf{x}'_j \mid \mathbf{x})}
\cdot
\frac{w_{\text{pos}}(m \mid \mathbf{x}'_j)}
     {w_{\text{pos}}(m \mid \mathbf{x})},
\end{equation}
with the acceptance rule of Eq.~\eqref{eq:mtm-acceptance} applied
to $W = \sum_j w_j^{\text{full}}$. Omitting the position factor
from Eq.~\eqref{eq:mh-exact} and from every weight in
Eq.~\eqref{eq:mtm-exact} yields exactly the simplified ratios
used by our implementation (Eqs.~\ref{eq:mh-acceptance}
and~\ref{eq:mtm-acceptance}).
Entropy-triggered skipping (\S\ref{sec:egps}) is the second
approximation: a block is skipped iff its maximum token entropy
is below $\theta$, where the conditionals are near-deterministic,
so the removed MCMC steps are those whose proposals would almost
surely reproduce the current tokens---skipping removes
near-trivial steps rather than discarding probability mass that
MH would have moved.

\paragraph{Experiment 1: acceptance-ratio level.}
We measure the change induced by restoring the position factor,
on the resampling steps actually executed by running EGPS
(Qwen2.5-Math-7B, all 500 MATH500 problems); skipped blocks
perform no resampling and involve no position factor, so this
measurement cleanly isolates the factor. For $K=1$ the omitted
factor's log,
$\Delta_{\text{pos}} = \log w_{\text{pos}}(m \mid \mathbf{x}')
- \log w_{\text{pos}}(m \mid \mathbf{x})$, is the entire effect
on the log ratio; for $K=5$, since $A_{\mathrm{MTM}}$ depends on
the sum of weights, we measure $\Delta A_{\mathrm{MTM}}$, the
change in $\log A_{\mathrm{MTM}}$ after restoring the factor to
every $w_j$. Table~\ref{tab:pos-factor} shows both concentrate
tightly at zero with no systematic direction;
$\Delta A_{\mathrm{MTM}}$ is tighter still because
$A_{\mathrm{MTM}} = \min(1, \cdot)$ is often saturated.

\paragraph{Experiment 2: output-distribution level.}
We compare full, skip-enabled EGPS against MH power
sampling~\citep{karan2025reasoning} with a two-sample
Kolmogorov--Smirnov test on the per-sequence mean log-likelihood
under the base model $p$ (Table~\ref{tab:ks-test}); this
end-to-end comparison covers the combined effect of both
approximations. The local variants are statistically
indistinguishable from MH power sampling; the global variant
shifts toward higher likelihood---a task-favorable
sharpening consistent with its stronger downstream accuracy
(\S\ref{sec:main_results})---while the reference row (MH power
sampling vs.\ plain sampling at $\alpha = 1$) confirms the test
has power to detect real distributional shifts.
Taken together, neither the omitted position factor nor
entropy-triggered skipping produces a detectable distributional
deviation for the local variants, and the global variant's
deviation is a beneficial sharpening.

\begin{table*}[!t]
\centering
\small
\begin{tabular}{@{}llcccc@{}}
\toprule
\textbf{Config} & \textbf{Quantity} & $|\Delta|{<}0.1$ & \textbf{Mean} & \textbf{Med.} & \textbf{Std} \\
\midrule
EGPS\textsubscript{L} (MH)  & $\Delta_{\text{pos}}$     & 66.5\% & 0.0004   & 0.000 & 0.302 \\
EGPS\textsubscript{G} (MTM) & $\Delta A_{\mathrm{MTM}}$ & 92.2\% & $-0.031$ & 0.000 & 1.036 \\
EGPS\textsubscript{L} (MTM) & $\Delta A_{\mathrm{MTM}}$ & 92.9\% & $-0.020$ & 0.000 & 0.224 \\
\bottomrule
\end{tabular}
\caption{Effect of restoring the position factor, measured on the
resampling steps executed by running EGPS (Qwen2.5-Math-7B,
MATH500; MH: $K{=}1$, MTM: $K{=}5$). For the MH
row, $1.4\%$ of accepted steps are reverse-unreachable:
after the move, the entropy-guided candidate region of the new
sequence no longer contains $m$, so
$w_{\text{pos}}(m \mid \mathbf{x}') = 0$ and
$\Delta_{\text{pos}} = -\infty$; these steps are excluded from
the statistics and reported separately.}
\label{tab:pos-factor}
\end{table*}

\begin{table}[!t]
\centering
\small
\begin{tabular}{@{}lccl@{}}
\toprule
\textbf{Comparison vs.\ MH PS} & \textbf{KS $D$} & KS $p$-value & \textbf{Verdict} \\
\midrule
EGPS\textsubscript{L} (MH)  & 0.040 & 0.82 & indist. \\
EGPS\textsubscript{L} (MTM) & 0.052 & 0.51 & indist. \\
EGPS\textsubscript{G} (MTM) & 0.146 & $<10^{-4}$ & higher lik. \\
\midrule
MH PS vs.\ base ($\alpha{=}1$) & 0.126 & $<10^{-3}$ & distinct \\
\bottomrule
\end{tabular}
\caption{Two-sample KS test on per-sequence mean log-likelihood
under the base model (Qwen2.5-Math-7B, MATH500). Local EGPS
variants are statistically indistinguishable from MH power
sampling; the global variant shifts toward higher likelihood. The
last row confirms the test detects real distributional shifts.}
\label{tab:ks-test}
\end{table}

\section{Derivation of the I-MTM Acceptance Ratio}
\label{app:imtm}

Our multiple-proposal mechanism is independent multiple-try
Metropolis (I-MTM)~\citep{martino2018review};
Eqs.~\eqref{eq:mtm-weight} and~\eqref{eq:mtm-acceptance} are the
standard I-MTM weight and acceptance ratio, Eqs.~(18)--(19) of
that survey (the I-MTM algorithm in its Table~6). Unlike standard
MTM, I-MTM does not require drawing $K{-}1$ auxiliary reference
points to balance the current state, which makes it simpler and
computationally more efficient.

Martino's acceptance ratio is expressed via the unnormalized
importance weight
$\Omega(\mathbf{z}) = \tilde{p}(\mathbf{z}) / q(\mathbf{z})$,
where $q$ denotes suffix regeneration from the prefix
$x_{<m}$ shared by the current state and all candidates: its
numerator is the sum of the weights over all $K$ candidates, and
its denominator is the current-state term $\Omega(\mathbf{x})$
plus the weights of the remaining $K{-}1$ candidates,
\begin{equation}\label{eq:imtm-martino}
A = \min\!\left(1,\;
\frac{\sum_{n=1}^{K} \Omega(\mathbf{x}'_n)}
     {\Omega(\mathbf{x}) + \sum_{n \ne j^{*}} \Omega(\mathbf{x}'_n)}
\right).
\end{equation}
We define each candidate weight as its ratio against the current
state, which is exactly Eq.~\eqref{eq:mtm-weight}:
\begin{equation}
w_j = \frac{\Omega(\mathbf{x}'_j)}{\Omega(\mathbf{x})}
    = \frac{\tilde{p}(\mathbf{x}'_j)\, q(\mathbf{x} \mid \mathbf{x}'_j)}
           {\tilde{p}(\mathbf{x})\, q(\mathbf{x}'_j \mid \mathbf{x})}.
\end{equation}
Dividing the numerator and denominator of
Eq.~\eqref{eq:imtm-martino} by $\Omega(\mathbf{x})$: the
numerator becomes $W = \sum_n w_n$, the current-state term
becomes $1$, and the remaining candidates contribute
$W - w_{j^{*}}$, yielding Eq.~\eqref{eq:mtm-acceptance}:
\begin{equation}
A_{\mathrm{MTM}} = \min\!\left(1,\; \frac{W}{W - w_{j^{*}} + 1}\right).
\end{equation}
Hence Eq.~\eqref{eq:mtm-acceptance} is the equivalent form of the
I-MTM acceptance ratio rather than a heuristic construction; the
$+1$ is the current-state term $\Omega(\mathbf{x})$ after
normalization, not an added constant. The selected candidate
$\mathbf{x}'_{j^{*}}$ is drawn with probability proportional to
$w_j$---the standard weighted selection of I-MTM (PMF
$\bar{w}_n = w_n / \sum_i w_i$). At $K = 1$ the denominator is
$w_1 - w_1 + 1 = 1$, recovering the standard MH acceptance
$\min(1, w_1)$. Thus the MTM mechanism introduces no
approximation beyond those analyzed in
Appendix~\ref{app:exactness}.

\section{Complexity Derivation}
\label{app:complexity}

\paragraph{MH power sampling baseline.}
With sequence length $T$, block size $B$, and $T/B$ blocks, the
MH sampler of \citet{karan2025reasoning} draws, at the $k$-th
block, a resampling start uniformly from the prefix of length
$\approx kB$, regenerating $\approx kB/2$ tokens per step in
expectation. Summing over blocks and multiplying by the
$N_{\mathrm{MCMC}}$ steps per block:
\begin{equation}
E_{\mathrm{MH}} \approx N_{\mathrm{MCMC}} \sum_{k=1}^{T/B}
\frac{kB}{2} \approx \frac{N_{\mathrm{MCMC}}\, T^{2}}{4B},
\end{equation}
quadratic in $T$.

\paragraph{EGPS.}
\begin{figure*}[!t]
\begin{equation}\label{eq:cost-local-full}
E_{\mathrm{local}} \le (1-s)\, K\, N_{\mathrm{MCMC}}
\cdot \frac{T}{B} \cdot B = (1-s)\, K\, N_{\mathrm{MCMC}}\, T
\end{equation}
\end{figure*}
The local mode restricts each rewrite to the current block, so
the regenerated length is at most $B$, independent of $k$. With
$K$ proposals per step ($K{=}1$: MH move; $K{>}1$: MTM move) and
block skip rate $s$, the total cost is bounded as in
Eq.~\eqref{eq:cost-local-full}, so $B$ cancels and the
complexity in $T$ drops from $O(T^{2}/B)$
to $O(KT)$---the source of EGPS\textsubscript{L}'s speedup. The
global mode allows the start anywhere in the prefix, so the worst
case remains $O(KT^{2}/B)$; but entropy weighting concentrates
resample points near the triggering block, so the empirical cost
is far lower (\S\ref{sec:ablation}). Larger $K$ raises the
nominal cost linearly, but our vLLM implementation batches the
$K$ proposals, so the added wall-clock latency is small.

\section{Component-Isolating Ablations}
\label{app:ablation-full}

We isolate each design choice under a matched compute budget
(MATH500, Qwen2.5-Math-7B, A100 40G; mean$\pm$std over 8 seeds).
The reference is our reproduced MH Power Sampling (global range,
no skipping, uniform position), $0.721 \pm 0.010$. Throughout,
$K{=}1$ denotes an MH move and $K{>}1$ an MTM move.

\paragraph{Tuning protocol.}
We first fix the MTM configuration on the uniform-position
variant (EGPS\textsubscript{L}, random position;
Tables~\ref{tab:ablation-k} and~\ref{tab:ablation-depth}):
$K{=}5$, $10$ MCMC steps, weighted selection. Tuning on the
uniform-position variant means $K$, the step count, and the
selection rule are chosen independently of the position-selection
strategy, so that when entropy-weighted position selection is
introduced afterwards, its own contribution can be isolated
cleanly. We then fix this configuration and vary one component at
a time (Tables~\ref{tab:ablation-runway}
and~\ref{tab:ablation-components}).

\paragraph{Findings.}
\begin{itemize}[topsep=2pt, itemsep=2pt, parsep=0pt, leftmargin=1.2em]
\item \textbf{MH vs.\ MTM} (Table~\ref{tab:ablation-components},
$K{=}1$ vs.\ $K{=}5$): the multi-proposal step is better for
$\theta \le 0.4$ (e.g., $.735$ vs.\ $.716$ at $\theta{=}0.1$).
\item \textbf{Entropy-weighted vs.\ uniform position}
(EGPS\textsubscript{L} vs.\ random pos.): entropy weighting wins
for $\theta \le 0.4$ ($.735/.738/.730$ vs.\ $.723/.721/.709$);
both rows use block skipping, so the gap comes from position
weighting alone.
\item \textbf{More steps instead of MTM} ($K{=}1$, $50$ steps):
reaches $.751 \pm .006$ at $\theta{=}0.1$ but requires $50$
serial MCMC steps; $K{=}5$ MTM attains comparable accuracy in
$10$ serial steps.
\item \textbf{Runway} (Table~\ref{tab:ablation-runway}): locking
the start onto high-entropy positions (runway $=0$) is best, with
$\delta \le 8$ nearly identical; spreading onto low-entropy
positions degrades accuracy (e.g., $.724$ at $16$)---high-entropy
positions are the critical resample locations.
\item \textbf{Weighted vs.\ greedy (best-of-$K$) selection}
(Tables~\ref{tab:ablation-k}
and~\ref{tab:ablation-components}): weighted matches or beats
greedy; greedy only catches up at $K{=}15$, so always taking the
best proposal is not optimal.
\item \textbf{Global vs.\ local}: EGPS\textsubscript{G} is
uniformly more accurate at the cost of the wall-clock speedup
dropping from ${\sim}3.5\times$ to ${\sim}1.1\times$ (measured
separately, \S\ref{sec:ablation}), so we provide both variants.
\end{itemize}

\begin{table*}[!t]
\centering
\footnotesize
\begin{tabular}{@{}lccccc@{}}
\toprule
$K$ & 1 & 3 & 5 & 10 & 15 \\
\midrule
Total cand. & 10 & 30 & 50 & 100 & 150 \\
Weighted & .705$\pm$.015 & .716$\pm$.012 & .723$\pm$.011 & .712$\pm$.012 & .717$\pm$.013 \\
Greedy   & .710$\pm$.008 & .707$\pm$.010 & .712$\pm$.015 & .709$\pm$.009 & .727$\pm$.007 \\
\bottomrule
\end{tabular}
\caption{Proposal count $K$ and selection rule on
EGPS\textsubscript{L} (random position), step $=10$,
$\theta{=}0.1$; MATH500, 8 seeds. Weighted at $K{=}5$ is the best
trade-off; greedy (best-of-$K$) only catches up at $K{=}15$.}
\label{tab:ablation-k}
\end{table*}

\begin{table*}[!t]
\centering
\footnotesize
\begin{tabular}{@{}lcccc@{}}
\toprule
Steps & 10 & 5 & 15 & 20 \\
\midrule
$K$ & 5 & 10 & 5 & 5 \\
Total cand. & 50 & 50 & 45 & 50 \\
Serial calls & 10 & 5 & 15 & 20 \\
Accuracy & .723$\pm$.011 & .705$\pm$.019 & .723$\pm$.013 & .728$\pm$.008 \\
\bottomrule
\end{tabular}
\caption{MCMC depth on EGPS\textsubscript{L} (random position),
weighted, $\theta{=}0.1$; MATH500, 8 seeds. At matched candidate
budget, accuracy barely improves beyond $10$ steps while time
grows proportionally; $K{=}10$/step $=5$ (same $50$ candidates,
half the depth) is clearly worse---deeper serial exploration
beats wider parallel proposals at fixed budget.}
\label{tab:ablation-depth}
\end{table*}

\begin{table*}[!t]
\centering
\footnotesize
\begin{tabular}{@{}lccccc@{}}
\toprule
Runway $\delta$ & 0 & 4 & 8 & 16 & 32 \\
\midrule
Accuracy & .739$\pm$.012 & .739$\pm$.012 & .738$\pm$.012 & .724$\pm$.015 & .732$\pm$.009 \\
Accept.\ (\%) & 82.7 & 82.4 & 82.1 & 82.0 & 81.5 \\
Gen.\ tokens & 17153 & 17155 & 17283 & 17304 & 17800 \\
\bottomrule
\end{tabular}
\caption{Runway sensitivity on full EGPS\textsubscript{L}
(entropy-weighted position, $K{=}5$, step $=10$,
$\theta{=}0.2$); MATH500, 8 seeds. Locking to high-entropy
positions (runway $=0$) is best, with $\delta \le 8$ nearly
identical; spreading further can hurt (e.g., runway $=16$).}
\label{tab:ablation-runway}
\end{table*}

\begin{table*}[!t]
\centering
\footnotesize
\setlength{\tabcolsep}{3pt}
\begin{tabular}{@{}lcccc@{}}
\toprule
Method & $\theta{=}0.1$ & $\theta{=}0.2$ & $\theta{=}0.4$ & $\theta{=}0.8$ \\
\midrule
L (random pos.) & .723$\pm$.011 & .721$\pm$.017 & .709$\pm$.005 & .707$\pm$.013 \\
L ($K{=}5$, MTM) & .735$\pm$.009 & .738$\pm$.012 & .730$\pm$.006 & .698$\pm$.011 \\
L ($K{=}1$, MH) & .716$\pm$.016 & .708$\pm$.010 & .713$\pm$.011 & .704$\pm$.013 \\
L ($K{=}5$, greedy) & .734$\pm$.013 & .734$\pm$.017 & .735$\pm$.013 & .703$\pm$.008 \\
G ($K{=}5$, MTM) & .746$\pm$.009 & .744$\pm$.009 & .744$\pm$.007 & .710$\pm$.010 \\
\bottomrule
\end{tabular}
\caption{Component effects at the fixed base configuration
($K{=}5$, step $=10$, weighted, runway $=8$) across entropy
thresholds $\theta$; MATH500, Qwen2.5-Math-7B, 8 seeds. L/G
abbreviate EGPS\textsubscript{L}/EGPS\textsubscript{G}. All rows
use entropy-triggered block skipping; ``random pos.'' uses
uniform intra-block position selection. Reference: reproduced MH
Power Sampling $= 0.721 \pm 0.010$ (constant over $\theta$).}
\label{tab:ablation-components}
\end{table*}

\section{Statistical Reliability over 8 Seeds}
\label{app:significance}

Table~\ref{tab:main} reports single-seed best accuracy following
the protocol of \citet{karan2025reasoning}. Here we rerun EGPS
and our reproduced MH Power Sampling baseline over 8 seeds
(0--7) and report mean$\pm$std, the gain $\Delta_{\text{acc}}$ over the
baseline with a two-sided paired $t$-test, and the 8-seed
wall-clock speedup (Table~\ref{tab:significance}). Within the
recommended range $\theta \le 0.4$:
\begin{itemize}[topsep=2pt, itemsep=2pt, parsep=0pt, leftmargin=1.2em]
\item \textbf{MATH500: significant gains for both variants.}
EGPS\textsubscript{G}: $+2.3$ to $+2.5$ points, significant at the $1\%$ level at
$1.1$--$1.2\times$ speedup; EGPS\textsubscript{L}: $+0.9$ to
$+1.7$ points, significant at the $5\%$ level at $3.5$--$3.8\times$. G gives the
larger accuracy gain, L the larger speedup.
\item \textbf{HumanEval / GPQA: on par, large speedup.} All mean
differences are positive but not significant; par accuracy is
the design goal, traded for $4.8$--$9.8\times$
(EGPS\textsubscript{L}) and $1.1$--$5.4\times$
(EGPS\textsubscript{G}) speedup.
\item \textbf{GPQA non-significance is a variance effect.} The
baseline fluctuates by $\pm 0.041$ across seeds, masking
moderate gains, while EGPS is more stable
($\pm 0.018$--$0.033$).
\item \textbf{Accuracy drops only at the non-recommended
$\theta{=}0.8$ on MATH500}, whose block-entropy mass peaks at
$0.6$--$0.75$ (Fig.~\ref{fig:entropy}c), so this threshold skips
critical blocks; HumanEval and GPQA peak outside this range and
do not drop---conversely confirming that skipping high-entropy
blocks is what costs accuracy.
\end{itemize}
Overall, the gains are stable and consistent with our claim of speedup with accuracy that improves or stays on par:
within $\theta \le 0.4$, speedup spans $1.1$--$9.8\times$ across
variants and datasets (up to $12.6\times$ at higher $\theta$).

\begin{table*}[!t]
\centering
\small
\begin{tabular*}{0.75\textwidth}{@{\extracolsep{\fill}}llccc@{}}
\toprule
Method & $\theta$ & Accuracy & $\Delta_{\text{acc}}$ & Speedup \\
\midrule
\multicolumn{5}{@{}l}{\textbf{MATH500} --- baseline $0.721\pm0.010$, $1\times$} \\
EGPS\textsubscript{G} & 0.1 & 0.746$\pm$0.009 & $+0.025^{**}$ & 1.1$\times$ \\
EGPS\textsubscript{G} & 0.2 & 0.744$\pm$0.009 & $+0.023^{**}$ & 1.1$\times$ \\
EGPS\textsubscript{G} & 0.4 & 0.744$\pm$0.007 & $+0.023^{**}$ & 1.2$\times$ \\
EGPS\textsubscript{G} & 0.8$^{\dagger}$ & 0.710$\pm$0.010 & $-0.011$ & 4.5$\times$ \\
EGPS\textsubscript{L} & 0.1 & 0.735$\pm$0.009 & $+0.015^{**}$ & 3.5$\times$ \\
EGPS\textsubscript{L} & 0.2 & 0.738$\pm$0.013 & $+0.017^{*}$ & 3.6$\times$ \\
EGPS\textsubscript{L} & 0.4 & 0.730$\pm$0.006 & $+0.009^{*}$ & 3.8$\times$ \\
EGPS\textsubscript{L} & 0.8$^{\dagger}$ & 0.698$\pm$0.012 & $-0.023^{***}$ & 8.5$\times$ \\
\midrule
\multicolumn{5}{@{}l}{\textbf{HumanEval} --- baseline $0.551\pm0.023$, $1\times$} \\
EGPS\textsubscript{G} & 0.1 & 0.564$\pm$0.013 & $+0.013$ & 4.0$\times$ \\
EGPS\textsubscript{G} & 0.2 & 0.557$\pm$0.026 & $+0.006$ & 4.5$\times$ \\
EGPS\textsubscript{G} & 0.4 & 0.573$\pm$0.020 & $+0.022$ & 5.4$\times$ \\
EGPS\textsubscript{G} & 0.8 & 0.579$\pm$0.025 & $+0.027^{*}$ & 9.5$\times$ \\
EGPS\textsubscript{L} & 0.1 & 0.566$\pm$0.026 & $+0.015$ & 9.3$\times$ \\
EGPS\textsubscript{L} & 0.2 & 0.568$\pm$0.021 & $+0.017$ & 9.6$\times$ \\
EGPS\textsubscript{L} & 0.4 & 0.555$\pm$0.018 & $+0.004$ & 9.8$\times$ \\
EGPS\textsubscript{L} & 0.8 & 0.568$\pm$0.025 & $+0.017$ & 12.6$\times$ \\
\midrule
\multicolumn{5}{@{}l}{\textbf{GPQA} --- baseline $0.339\pm0.041$, $1\times$} \\
EGPS\textsubscript{G} & 0.1 & 0.349$\pm$0.018 & $+0.010$ & 1.1$\times$ \\
EGPS\textsubscript{G} & 0.2 & 0.354$\pm$0.031 & $+0.015$ & 1.2$\times$ \\
EGPS\textsubscript{G} & 0.4 & 0.357$\pm$0.031 & $+0.018$ & 1.3$\times$ \\
EGPS\textsubscript{G} & 0.8 & 0.360$\pm$0.021 & $+0.021$ & 2.8$\times$ \\
EGPS\textsubscript{L} & 0.1 & 0.340$\pm$0.025 & $+0.001$ & 4.9$\times$ \\
EGPS\textsubscript{L} & 0.2 & 0.343$\pm$0.033 & $+0.004$ & 4.8$\times$ \\
EGPS\textsubscript{L} & 0.4 & 0.343$\pm$0.029 & $+0.004$ & 5.2$\times$ \\
EGPS\textsubscript{L} & 0.8 & 0.363$\pm$0.030 & $+0.024$ & 6.1$\times$ \\
\bottomrule
\end{tabular*}
\caption{Statistical reliability of EGPS vs.\ the reproduced MH
Power Sampling baseline over 8 seeds (Qwen2.5-Math-7B). Accuracy
is the 8-seed mean$\pm$std; $\Delta_{\text{acc}}$ is the gain over the
baseline with a two-sided paired $t$-test
($^{*}$: significant at the $5\%$ level; $^{**}$: $1\%$; $^{***}$: $0.1\%$); speedup is the
8-seed wall-clock ratio. $^{\dagger}$On MATH500,
$\theta{=}0.8$ is a non-recommended high-threshold setting
(see text). Recommended range: $\theta \le 0.4$.}
\label{tab:significance}
\end{table*}

\section{Mechanism-Level Efficiency Metrics}
\label{app:efficiency}

Wall-clock speedup is hardware-dependent, so we complement it
with hardware-independent, mechanism-level efficiency metrics
(Table~\ref{tab:efficiency}) plus a measured GPU peak-memory
figure. Our implementation decodes one token per forward pass
and counts all $K$ MTM candidate chains, so ``generated
tokens'' equals the total number of forward passes---the most
direct hardware-independent measure of compute.

\paragraph{Compute decreases monotonically with $\theta$.}
As $\theta$ rises, the trigger rate, the number of refined
blocks, the forward calls, and the generated-to-output token
ratio all fall while the final answer length is essentially
unchanged, cutting total forward compute by roughly $3$--$5\times$
at matched length between $\theta{=}0.1$ and $\theta{=}0.8$.
Both variants show the same trend, indicating that the speedup
comes from genuinely less refinement rather than implementation
or hardware effects.

\paragraph{The gate is deterministic.}
A block is skipped iff its maximum token entropy is below
$\theta$, so no false positives or negatives arise from the
trigger itself; that the selected positions are the
answer-changing ones is validated separately by the
position-selection ablation (Appendix~\ref{app:ablation-full}).

\paragraph{Peak memory is independent of $\theta$.}
Since EGPS changes neither the model weights, the KV-cache
budget, nor $K$, peak GPU memory stays at $36.7$\,GB (A100 40G)
across all $\theta$ and both variants.

\begin{table*}[!t]
\centering
\footnotesize
\setlength{\tabcolsep}{3pt}
\begin{tabular}{@{}lccccccccc@{}}
\toprule
Variant & $\theta$ & Trig.\ rate & Accept.\ (\%) & Refined & Skipped & Skip/Refine & Gen/Out ($\times$) & Gen.\ tokens & Out.\ tokens \\
\midrule
EGPS\textsubscript{G} (MTM) & 0.1 & .779 & 61.8 & 15375 & 4358  & 0.28 & 89.0 & 63091 & 708.6 \\
EGPS\textsubscript{G} (MTM) & 0.2 & .753 & 61.4 & 14543 & 4784  & 0.33 & 84.2 & 59332 & 704.9 \\
EGPS\textsubscript{G} (MTM) & 0.4 & .702 & 60.3 & 13341 & 5668  & 0.42 & 76.7 & 53875 & 702.2 \\
EGPS\textsubscript{G} (MTM) & 0.8 & .255 & 57.4 & 4115  & 12049 & 2.93 & 19.5 & 12813 & 655.7 \\
\midrule
EGPS\textsubscript{L} (MTM) & 0.1 & .805 & 82.6 & 13955 & 3385  & 0.24 & 26.6 & 17743 & 666.2 \\
EGPS\textsubscript{L} (MTM) & 0.2 & .777 & 82.1 & 13330 & 3837  & 0.29 & 26.0 & 17283 & 664.2 \\
EGPS\textsubscript{L} (MTM) & 0.4 & .727 & 80.2 & 12287 & 4621  & 0.38 & 24.6 & 16253 & 659.7 \\
EGPS\textsubscript{L} (MTM) & 0.8 & .255 & 68.6 & 3940  & 11526 & 2.93 & 9.4  & 5996  & 635.5 \\
\bottomrule
\end{tabular}
\caption{Mechanism-level efficiency metrics of EGPS
(hardware-independent). Qwen2.5-Math-7B, MATH500, $K{=}5$, 8
seeds, A100 40G. ``Accept.'' is the MCMC acceptance rate;
``Gen.\ tokens'' equals the total number of forward passes
(one token per pass, all $K$ candidate chains counted);
``Refined''/``Skipped'' count blocks; a smaller Gen/Out ratio
means fewer generated tokens per final answer token. Generated
and output tokens are per-problem averages.}
\label{tab:efficiency}
\end{table*}

\section{Per-$\theta$ Numerical Results}
\label{app:theta-numbers}

Table~\ref{tab:theta-numbers} lists the full numerical results
behind Figure~\ref{fig:ablation}. Accuracy is the best value
from a single seed (matching the protocol of
Table~\ref{tab:main}); speedup is averaged over 8 seeds and
measured against our vLLM reproduction of MH power sampling, so
it reflects the algorithm rather than framework factors.

\begin{table*}[!t]
\centering
\footnotesize
\begin{tabular}{@{}llcccccccc@{}}
\toprule
& & \multicolumn{4}{c}{Accuracy} & \multicolumn{4}{c}{Speedup ($\times$)} \\
\cmidrule(lr){3-6} \cmidrule(l){7-10}
Dataset & Method & $\theta{=}.1$ & $.2$ & $.4$ & $.8$ & $\theta{=}.1$ & $.2$ & $.4$ & $.8$ \\
\midrule
MATH500 & EGPS\textsubscript{G} (MTM) & .756 & .758 & .756 & .714 & 1.1 & 1.1 & 1.2 & 4.5 \\
MATH500 & EGPS\textsubscript{L} (MTM) & .746 & .754 & .742 & .708 & 3.5 & 3.6 & 3.8 & 8.5 \\
MATH500 & EGPS\textsubscript{L} (MH) & .742 & .724 & .724 & .726 & 3.8 & 3.9 & 4.1 & 8.7 \\
MATH500 & EGPS\textsubscript{L} (MTM, rand.) & .738 & .746 & .716 & .728 & 4.0 & 4.2 & 4.4 & 9.5 \\
\midrule
HumanEval & EGPS\textsubscript{G} (MTM) & .585 & .604 & .598 & .604 & 4.0 & 4.5 & 5.4 & 9.5 \\
HumanEval & EGPS\textsubscript{L} (MTM) & .622 & .591 & .585 & .604 & 9.3 & 9.6 & 9.8 & 12.6 \\
HumanEval & EGPS\textsubscript{L} (MH) & .585 & .573 & .567 & .573 & 10.8 & 11.1 & 11.6 & 14.8 \\
HumanEval & EGPS\textsubscript{L} (MTM, rand.) & .579 & .579 & .579 & .610 & 10.2 & 10.8 & 10.5 & 15.0 \\
\midrule
GPQA & EGPS\textsubscript{G} (MTM) & .343 & .404 & .424 & .359 & 1.09 & 1.19 & 1.32 & 2.82 \\
GPQA & EGPS\textsubscript{L} (MTM) & .354 & .384 & .379 & .384 & 4.85 & 4.80 & 5.15 & 6.05 \\
GPQA & EGPS\textsubscript{L} (MH) & .399 & .389 & .364 & .343 & 5.2 & 5.3 & 5.5 & 7.93 \\
GPQA & EGPS\textsubscript{L} (MTM, rand.) & .384 & .333 & .349 & .379 & 5.85 & 5.64 & 5.92 & 9.2 \\
\bottomrule
\end{tabular}
\caption{Full per-$\theta$ numerical results behind
Figure~\ref{fig:ablation} (Qwen2.5-Math-7B). Accuracy is the
single-seed best; speedup is the 8-seed average relative to the
vLLM reproduction of MH Power Sampling.}
\label{tab:theta-numbers}
\end{table*}

\section{Prompts and Decoding Configuration}
\label{app:prompts}

Apart from EGPS's three new parameters---the proposal count $K$,
the runway $\delta$, and the entropy threshold $\theta$---the
prompt templates, the zero-shot chain-of-thought setup, and all
decoding hyperparameters match MH Power
Sampling~\citep{karan2025reasoning}; the shared hyperparameter
values are listed in \S\ref{sec:setup}.
Table~\ref{tab:prompts} lists the zero-shot prompts. For the
base models (Qwen2.5-Math-7B, Qwen2.5-7B) the prompt is fed
directly as plain text; for the instruction-tuned
Phi-3.5-mini-instruct the same content is wrapped as a user
message in the model's chat template with the generation prompt
appended.

\begin{table*}[!t]
\centering
\small
\begin{tabular}{@{}lp{0.78\textwidth}@{}}
\toprule
\textbf{Benchmark} & \textbf{Zero-shot prompt} \\
\midrule
MATH500 & \ttfamily Can you solve the following math problem?
\{question\} Please reason step by step, and put your final
answer within \textbackslash{}boxed\{\}. \\
\addlinespace
HumanEval & The unmodified dataset prompt (function signature
and docstring), used directly as the completion prefix with no
additional instruction. \\
\addlinespace
GPQA Diamond & \ttfamily Answer the following multiple choice
question. The last line of your response should be of the
following format: '\textbackslash{}boxed\{\$LETTER\}' (without
quotes) where LETTER is one of ABCD (ex.
'\textbackslash{}boxed\{A\}'). Think step by step before
answering.\newline
\{Question\}\newline
A) \{A\}\newline
B) \{B\}\newline
C) \{C\}\newline
D) \{D\} \\
\bottomrule
\end{tabular}
\caption{Zero-shot prompt templates, identical to MH Power
Sampling~\citep{karan2025reasoning}. Typewriter text is the
literal template; \{$\cdot$\} marks the slot filled per
problem.}
\label{tab:prompts}
\end{table*}

\section{Comparison with Entropy-Based Adaptive Sampling}
\label{app:adaptive}

\paragraph{Reproduction setup.}
\citet{abdulloh2025uncertainty} is the closest work to ours: it likewise
uses entropy to allocate the MCMC budget, but does so by adaptively adjusting the block size (finer blocks in uncertain
regions), whereas EGPS keeps the block size fixed, skips low-entropy
blocks outright, and concentrates resampling on high-entropy positions.
The method is not open-sourced and its original implementation is in
PyTorch; to obtain a framework-matched comparison, we reproduced it on
vLLM following the description in the paper, denoted
Adaptive$^{\dagger}$ (single seed).
Parameters follow the best configuration reported in that paper:
$b_{\min}{=}48$, $b_{\max}{=}384$, uncertainty threshold $1.8$, max
aggregation, and $N_{\mathrm{MCMC}}{=}10$; $\alpha{=}4$,
$\tau{=}0.25$, and a maximum generation length of $3072$ are shared with
MH Power Sampling. The paper states that this threshold is calibrated on
a held-out set for a target compute budget, and we adopt its reported
value. We use base models, and GPQA answers are graded with the grader
from the original repository of \citet{karan2025reasoning}. The
efficiency comparison is run on an A100-40G GPU on GPQA Diamond, with
all three methods measured under the same hardware and the same
framework.

\paragraph{Results.}
Accuracy is broadly comparable between the two methods
(Table~\ref{tab:adaptive-acc}). On the two models for which the original
paper reports numbers, the two trade wins: on Qwen2.5-7B, Adaptive is
slightly ahead on MATH500 and about $3$ points behind on GPQA, while on
Phi-3.5-mini-instruct it is stronger on HumanEval and GPQA---consistent
with the observation in \S\ref{sec:main_results} that instruction tuning
sharpens the output distribution and leaves less redirectable
high-entropy mass. Differences on GPQA should not be over-interpreted:
the baseline's own cross-seed standard deviation on this benchmark
reaches $\pm 0.041$ (Appendix~\ref{app:significance}).

Efficiency differs substantially (Table~\ref{tab:adaptive-eff}).
EGPS\textsubscript{L} is $5.0$--$8.2\times$ faster than
Adaptive$^{\dagger}$ across the three models; taking the same-framework
MH Power Sampling baseline on Qwen2.5-Math-7B as an absolute reference
($356.4$ seconds per problem, 8 seeds), Adaptive$^{\dagger}$ runs at
$0.96\times$---that is, no speedup at all---whereas
EGPS\textsubscript{L} reaches $4.83\times$. The reason lies in the
different form of intervention: the adaptive strategy lowers the mean
committed block from $192$ to roughly $120$ tokens, and the resulting
increase in block count raises the number of MCMC rounds accordingly, so
it reallocates computational granularity without reducing the
total budget. EGPS instead removes computation: low-entropy
blocks are skipped entirely and each move is confined to the current
block, so cost scales with entropy mass rather than sequence length
(\S\ref{sec:egps}).

\begin{table*}[!t]
\centering
\small
\begin{tabular}{@{}lcccc@{}}
\toprule
\textbf{Benchmark} & \textbf{Adaptive (reported)} & \textbf{Adaptive$^{\dagger}$ (ours)} & \textbf{EGPS\textsubscript{G}} & \textbf{EGPS\textsubscript{L}} \\
\midrule
\multicolumn{5}{@{}l}{\textbf{Qwen2.5-Math-7B}} \\
MATH500   & ---   & 0.720 & 0.758 & 0.754 \\
HumanEval & ---   & 0.524 & 0.604 & 0.622 \\
GPQA      & ---   & 0.337 & 0.424 & 0.384 \\
\midrule
\multicolumn{5}{@{}l}{\textbf{Qwen2.5-7B}} \\
MATH500   & 0.714 & 0.688 & 0.710 & 0.712 \\
HumanEval & 0.645 & 0.634 & 0.610 & 0.646 \\
GPQA      & 0.326 & 0.293 & 0.354 & 0.359 \\
\midrule
\multicolumn{5}{@{}l}{\textbf{Phi-3.5-mini-instruct}} \\
MATH500   & 0.521 & 0.504 & 0.520 & 0.504 \\
HumanEval & 0.748 & 0.665 & 0.677 & 0.695 \\
GPQA      & 0.372 & 0.328 & 0.343 & 0.338 \\
\bottomrule
\end{tabular}
\caption{Accuracy comparison (pass@1). All numbers are single-seed
results, with both methods evaluated at their own best configuration:
EGPS entries are the best over the $\theta$ sweep (as in
Table~\ref{tab:main}), and Adaptive$^{\dagger}$ uses the best
configuration reported in the original paper. On
Qwen2.5-Math-7B$\times$GPQA, 5 problems whose sequences reached the
model's 4096-token context limit crashed the inference engine and
produced no output; they are excluded ($n{=}193$). The original paper
does not report Qwen2.5-Math-7B, so only reproduced numbers are
available for that model. Our reproduction falls $1$--$8$ points below
the reported numbers; without a reference implementation it may not
capture every implementation detail. Eight-seed means for
Qwen2.5-Math-7B are given in Appendix~\ref{app:significance}.}
\label{tab:adaptive-acc}
\end{table*}

\begin{table*}[!t]
\centering
\small
\begin{tabular}{@{}llccc@{}}
\toprule
\textbf{Model} & \textbf{Method} & \textbf{Sec./problem} & \textbf{Gen./out.\ tokens} & \textbf{vs.\ Adaptive$^{\dagger}$} \\
\midrule
\multirow{3}{*}{Qwen2.5-Math-7B}
 & Adaptive$^{\dagger}$                  & 370.8 & 38.1$\times$  & 1.00$\times$ \\
 & EGPS\textsubscript{G}, $\theta{=}0.1$ & 330.5 & 101.1$\times$ & 1.12$\times$ \\
 & EGPS\textsubscript{L}, $\theta{=}0.1$ & 73.8  & 22.6$\times$  & \textbf{5.03$\times$} \\
\midrule
\multirow{3}{*}{Qwen2.5-7B}
 & Adaptive$^{\dagger}$                  & 348.7 & 44.2$\times$  & 1.00$\times$ \\
 & EGPS\textsubscript{G}, $\theta{=}0.1$ & 200.1 & 64.3$\times$  & 1.74$\times$ \\
 & EGPS\textsubscript{L}, $\theta{=}0.1$ & 53.8  & 22.9$\times$  & \textbf{6.48$\times$} \\
\midrule
\multirow{3}{*}{Phi-3.5-mini-instruct}
 & Adaptive$^{\dagger}$                  & 202.5 & 50.3$\times$  & 1.00$\times$ \\
 & EGPS\textsubscript{G}, $\theta{=}0.1$ & 82.1  & 85.6$\times$  & 2.47$\times$ \\
 & EGPS\textsubscript{L}, $\theta{=}0.1$ & 24.6  & 23.6$\times$  & \textbf{8.23$\times$} \\
\bottomrule
\end{tabular}
\caption{Efficiency comparison on GPQA Diamond (A100-40G, vLLM).
Statistics are computed over all problems. The generated-to-output token
ratio is a full count for both methods: for EGPS it includes all $K$ MTM
candidate chains, and for Adaptive$^{\dagger}$ it includes tokens
discarded by lookahead decoding as well as rejected proposals. Because
the $K$ candidates are batched in parallel under vLLM, this ratio
upper-bounds EGPS's latency cost---which is why EGPS\textsubscript{G}
remains faster than Adaptive$^{\dagger}$ despite a ratio of
$101\times$.}
\label{tab:adaptive-eff}
\end{table*}

\end{document}